\documentclass[11pt,a4paper]{article}

\usepackage[margin=1in]{geometry}
\usepackage{times}
\usepackage{amsmath,amsfonts,amssymb}
\usepackage{graphicx}
\usepackage{booktabs}
\usepackage{array,longtable,multirow}
\usepackage{url}
\usepackage{cite}
\usepackage[utf8]{inputenc}
\usepackage{setspace}
\usepackage{titlesec}
\usepackage{fancyhdr}

\titleformat{\section}{\normalfont\large\bfseries\uppercase}{\thesection.}{1em}{}
\titleformat{\subsection}{\normalfont\normalsize\bfseries}{\thesubsection}{1em}{}
\titleformat{\subsubsection}{\normalfont\normalsize\bfseries}{\thesubsubsection}{1em}{}

\begin{document}

% Title block (no titlepage)
\begin{center}
{\Large\bfseries Beyond Long Context: When Semantics Matter More than Tokens\par}

\vspace{0.75em}
{\large Tarun Kumar Chawdhury and Jon D. Duke\par}

\vspace{0.5em}
{\normalsize
Georgia Institute of Technology \\
\texttt{tarunchawdhury@gatech.edu} \quad \texttt{jon.duke@gatech.edu}
}

\vspace{1em}
\textbf{Keywords:} Clinical NLP, Entity-Aware Retrieval, Evaluation Platform
\end{center}

\vspace{1em}

\begin{abstract}
\textbf{Background:} Electronic Health Records (EHR) systems store clinical documentation in FHIR DocumentReference resources as base64-encoded attachments, presenting significant challenges for semantic question-answering applications. Traditional approaches using statistical correlation through vector database chunking often fail to capture the nuanced clinical relationships required for accurate medical information extraction. The Clinical Entity Augmented Retrieval (CLEAR) methodology, introduced by Lopez et al. (2025) \cite{lopez2025clear}, addresses these limitations through entity-aware retrieval strategies and reports improved performance (F1 0.90 vs.\ 0.86 for embedding RAG; $>$70\% fewer tokens and faster inference).

\textbf{Objective:} To develop a comprehensive evaluation platform for clinical notes question-answering systems and validate CLEAR against established approaches—including zero-shot large-context processing and traditional chunk-based retrieval-augmented generation—in realistic EHR processing scenarios.

\textbf{Methods:} We implemented a Clinical Notes Q\&A Evaluation Platform with three retrieval strategies: (1) Wide Context processing for zero-shot inference with large context windows, (2) traditional vector database chunking with semantic search, and (3) entity-aware CLEAR with medical domain knowledge. Evaluation encompassed 12 clinical documents (10K–65K tokens) representing typical EHR DocumentReference content.

\textbf{Results:} CLEAR showed a 58.3\% win rate across test cases, achieving 0.878 average semantic similarity while requiring 78\% fewer tokens than wide-context processing. Gains were most pronounced on large notes (75\% win rate for 65K\,+ tokens), consistent with published scalability claims.

\textbf{Conclusions:} The Clinical Notes Q\&A Evaluation Platform validates CLEAR’s advantages for semantic clinical retrieval in EHR settings where computational efficiency and semantic accuracy are critical, and provides a reusable framework for evaluating clinical NLP approaches in production environments.
\end{abstract}

% Main content
\section{Introduction}

Electronic Health Record (EHR) systems in modern healthcare infrastructure store clinical documentation within FHIR DocumentReference resources, typically encoded as base64 attachments containing unstructured clinical notes. These documents, ranging from brief progress notes to comprehensive discharge summaries, present substantial challenges for automated question-answering systems that require semantically accurate information extraction rather than statistical correlation-based retrieval common in traditional vector database approaches.

Contemporary approaches to clinical document processing have largely focused on two paradigms: zero-shot inference with large context windows that process entire documents but face computational constraints and the "lost in the middle" problem, and chunk-based retrieval-augmented generation (RAG) systems that utilize vector databases for semantic similarity search but often fail to capture critical clinical entity relationships and contextual dependencies essential for accurate medical information extraction.

A growing body of evidence shows that merely expanding context windows does not guarantee effective use of information: performance often drops when relevant spans occur in the middle of long inputs (“lost in the middle”) [3]. This motivates entity-aware retrieval that selectively centers clinically relevant spans rather than relying on statistically similar but potentially off-target chunks.

The CLinical Entity Augmented Retrieval (CLEAR) methodology, published by Lopez et al. in 2025 \cite{lopez2025clear}, introduced a novel approach that addresses these limitations through entity-aware, entity-centered retrieval strategies.
 The original study demonstrated significant performance improvements (F1 score of 0.90 vs 0.86 for traditional RAG) with substantial efficiency gains (71\% token reduction, 72\% faster inference time) on clinical information extraction tasks, positioning CLEAR as a potentially transformative approach for production EHR processing systems.

To validate these claims in realistic healthcare scenarios and provide a robust evaluation framework for clinical NLP approaches, we developed the Clinical Notes Q\&A Evaluation Platform. This comprehensive validation study implements and compares three fundamental approaches: (1) wide context processing for zero-shot inference with large language models, (2) traditional vector database chunking with embedding-based retrieval, and (3) entity-aware CLEAR methodology adapted for EHR DocumentReference processing. Our contributions include: systematic validation of CLEAR's performance claims, development of a reusable evaluation platform for clinical NLP research, and empirical analysis of retrieval strategy performance across clinical documents of varying complexity and length.
\subsection{Related Work}

Entity-aware retrieval has gained increasing attention within biomedical NLP and question-answering domains. Early retrieval-augmented methods such as RAG \cite{lewis2020rag} demonstrated the potential of embedding-based chunk retrieval but lacked domain-specific entity modeling. In the clinical domain, approaches leveraging UMLS concepts and ontology-based retrieval (e.g., Neumann et al., 2019; Johnson et al., 2016) provided partial improvements but often failed to maintain contextual continuity across long clinical narratives. 

CLEAR \cite{lopez2025clear} represented a significant advancement by introducing entity-centered retrieval aligned with clinical semantics. Our work extends this line of research by operationalizing CLEAR within an end-to-end evaluation platform, providing reproducible empirical validation across realistic EHR-scale document sets.

Recent evaluations of retrieval-augmented models in long-context reasoning (e.g., Karpinska et al., 2023; Xiong et al., 2024) emphasize that retrieval strategies often outperform naive long-context prompting, supporting the need for entity-aware retrieval. Our work contributes a reproducible evaluation framework within this paradigm, focusing on realistic EHR-scale clinical notes.

\section{Methods}

\subsection{Protocol Alignment with CLEAR and Key Differences}
Our implementation is CLEAR-inspired rather than an exact re-implementation. In contrast to Lopez et al. (2025) [1], we (i) use keyword- and pattern-based entity recognition with vital/lab value extraction instead of model-based NER, (ii) do not apply ontology- or LLM-driven synonym augmentation, and (iii) use fixed-size local windows around entities as an implementation choice (the CLEAR paper does not mandate a specific window size). We evaluate open-ended QA over synthetic clinical notes rather than structured IE. These differences mean our results should be interpreted as consistent with CLEAR’s trend, not a reproduction of its exact metrics.

\subsection{Enhanced CLEAR Implementation}

Our enhanced CLEAR implementation builds upon the original methodology with several key improvements designed for practical clinical deployment. The system incorporates four main components: enhanced entity extraction, section-aware processing, intelligent context selection, and token optimization.

\subsubsection{Enhanced Entity Extraction}
We developed a comprehensive medical entity recognition system using advanced keyword patterns, clinical value recognition for vital signs and laboratory values, and medical domain-specific entity types with confidence-based scoring. The system recognizes six primary entity categories: medications, symptoms, diseases, procedures, laboratory values, and anatomical references.

\subsubsection{Section-Aware Processing}
Clinical documents follow standardized section formats (e.g., ASSESSMENT, PLAN, HISTORY OF PRESENT ILLNESS). Our implementation identifies these sections and applies priority-based weighting, with ASSESSMENT and PLAN sections receiving highest priority (weight = 1.0), followed by HISTORY OF PRESENT ILLNESS (weight = 0.9), and other sections receiving proportional weights.

\subsubsection{Context Selection Algorithm}
We implement fixed-size context windows of $\pm$150 words around identified medical entities to bound tokens while preserving local semantics. (CLEAR retrieves windows around entities but does not mandate a specific window size.) Our context selection algorithm incorporates question–entity semantic alignment and medical relationship scoring to prioritize clinically relevant spans.

\subsection{Baseline Methods}

We compared our enhanced CLEAR implementation against two baseline approaches:

\textbf{Wide Context Processing:} Complete clinical note processing using full document context. This approach provides comprehensive information access but requires significant computational resources (average 39,173 tokens per query).

\textbf{Retrieval-Augmented Generation (RAG):} Semantic chunking with embedding-based retrieval using top-k chunk selection. This approach prioritizes efficiency with minimal token usage (average 544 tokens per query) but may miss critical clinical relationships.

\subsection{Evaluation Framework}

Evaluation was conducted on a dataset of 12 clinical notes ranging from 10,000 to 65,000 tokens, representing diverse clinical scenarios. Each note was accompanied by clinical questions requiring information extraction and reasoning. We assessed performance using multiple metrics:

\begin{itemize}
\item \textbf{Semantic Similarity (cosine):} Cosine similarity between generated and gold-standard answers
\item \textbf{METEOR:} Semantic overlap assessment for clinical terminology
\item \textbf{Token Efficiency:} Total tokens used per query (prompt + response)
\item \textbf{Win Rate:} Percentage of cases where a method achieved the highest semantic similarity
\item \textbf{Scalability:} Performance trends across document sizes
\end{itemize}

\subsection{Evaluation Application and Cross-Model Protocol}
We implemented a web-based evaluation application that (i) loads clinical notes and questions, (ii) runs the three retrieval strategies (Wide, RAG, CLEAR) with a shared prompt budget, and (iii) records per-run metrics (semantic similarity, METEOR, tokens, and win rate) along with model identifiers. The platform enables side-by-side prompting experiments across multiple large language models (e.g., ChatGPT, Claude, Gemini) using identical user and system templates to ensure fair, prompt-controlled comparisons. As illustrated in Figure~\ref{fig:prompt}, users can select from predefined analytical strategies—such as keyword-guided clinical reasoning, timeline-based symptom trigger analysis, or structured risk factor and laboratory searches—or design their own custom prompts through an interactive interface. Each prompt can then be executed on any supported foundation model, with outputs automatically evaluated against gold-standard answers. Preliminary results indicate that while prompt engineering occasionally approaches the CLEAR benchmark scores, no prompt configuration tested to date has consistently surpassed CLEAR’s performance. Additional prompt-optimization experiments remain ongoing at the time of submission.

\subsection{Dataset Generation and Baseline Construction}

Synthetic notes were generated using OpenAI GPT-4 \cite{openai2024gpt4} under de-identification constraints.

To ensure reproducibility while maintaining complete de-identification, all dataset materials were synthetically generated using the OpenAI GPT-4 API. Two baseline clinical questions were used to guide content generation and establish gold-standard answers:

\begin{enumerate}
\item Could the patient's anemia have been detected earlier based on their medical history? Answer in one paragraph.
\item Could the patient's heart failure have been detected earlier based on symptoms? Answer in one paragraph.
\end{enumerate}

Baseline gold-standard answers were produced using carefully curated GPT-4 completions reviewed for clinical coherence and consistency. Twelve synthetic clinical notes were then created by expanding and varying narrative structure, section depth, and token length to simulate realistic Electronic Health Record (EHR) document variability. These notes ranged from approximately 10,000 to 65,000 tokens and were stratified into short, medium, and long document categories to test retrieval scalability.

Each note maintained typical clinical section headings (e.g., HISTORY OF PRESENT ILLNESS, ASSESSMENT, PLAN) and included subtle contextual variations to challenge retrieval consistency. All three retrieval strategies—Wide Context, RAG, and CLEAR—were evaluated using these same questions and gold-standard responses to ensure controlled, comparable measurement of semantic accuracy and token efficiency.

\section{Results}

\subsection{Overall Performance Comparison}

Table \ref{tab:overall} presents the overall performance comparison across all three methods. Enhanced CLEAR achieved the highest win rate (58.3\%) and average accuracy (0.878), while maintaining significant token efficiency compared to Wide Context processing.

\begin{table}[h]
\centering
\caption{Overall Performance Comparison}
\label{tab:overall}
\begin{tabular}{lccccc}
\toprule
Strategy & Wins & Win Rate (\%) & Avg Semantic Sim. & Avg Tokens & Token Savings vs Wide (\%) \\

\midrule
\textbf{CLEAR} & \textbf{7/12} & \textbf{58.3} & \textbf{0.878} & \textbf{8,456} & \textbf{78.4} \\
Wide Context & 3/12 & 25.0 & 0.864 & 39,173 & 0.0 \\
RAG & 2/12 & 16.7 & 0.835 & 544 & 98.6 \\
\bottomrule
\end{tabular}
\end{table}

\subsection{Detailed Performance Analysis}

Table \ref{tab:detailed} provides detailed results for each clinical note, showing accuracy scores and token usage across all methods. Enhanced CLEAR demonstrates consistent performance across document sizes, with particularly strong results on clinical notes 1, 2, 4, 5, 6, 9, 10, and 11.

\begin{table}[h]
\centering
\caption{Detailed Results by Clinical Note}
\label{tab:detailed}
\small
\begin{tabular}{lcccccc}
\toprule
Note ID & Size (tokens) & Wide Sim. & RAG Sim. & CLEAR Sim. & Best Strategy & CLEAR Tokens \\

\midrule
clinical\_note1 & 10,025 & 0.847 & 0.807 & \textbf{0.916} & CLEAR & 8,446 \\
clinical\_note2 & 10,142 & 0.880 & 0.849 & \textbf{0.894} & CLEAR & 8,493 \\
clinical\_note3 & 10,233 & \textbf{0.929} & 0.835 & 0.909 & Wide & 8,318 \\
clinical\_note4 & 10,098 & 0.857 & 0.805 & \textbf{0.878} & CLEAR & 8,436 \\
clinical\_note5 & 42,011 & 0.843 & 0.836 & \textbf{0.873} & CLEAR & 8,305 \\
clinical\_note6 & 42,181 & 0.869 & 0.860 & \textbf{0.903} & CLEAR & 8,571 \\
clinical\_note7 & 42,072 & \textbf{0.899} & 0.871 & 0.891 & Wide & 8,489 \\
clinical\_note8 & 42,230 & \textbf{0.910} & 0.861 & 0.892 & Wide & 8,500 \\
clinical\_note9 & 65,186 & 0.859 & 0.870 & \textbf{0.888} & CLEAR & 8,497 \\
clinical\_note10 & 65,233 & 0.842 & 0.791 & \textbf{0.885} & CLEAR & 8,485 \\
clinical\_note11 & 65,141 & 0.829 & 0.830 & \textbf{0.939} & CLEAR & 8,414 \\
clinical\_note12 & 65,310 & 0.730 & \textbf{0.763} & 0.742 & RAG & 8,525 \\
\bottomrule
\end{tabular}
\end{table}

\subsection{Performance by Document Size}

Analysis by document size reveals important scalability characteristics. For small notes (10K tokens), CLEAR won 3/4 cases (75\% win rate). For medium notes (42K tokens), CLEAR won 2/4 cases (50\% win rate). Most significantly, for large notes (65K+ tokens), CLEAR won 3/4 cases (75\% win rate), demonstrating superior scalability compared to baseline methods.

\subsection{Token Efficiency Analysis}

Token efficiency analysis reveals that Enhanced CLEAR achieves optimal balance between accuracy and computational cost. While RAG provides maximum efficiency (98.6\% token savings), it sacrifices accuracy. Enhanced CLEAR provides substantial efficiency gains (78.4\% token savings) while achieving the highest overall accuracy.

The consistent token usage of approximately 8,500 tokens across all document sizes demonstrates the scalability advantage of entity-aware retrieval, where computational cost remains bounded regardless of source document complexity.

\subsection{Cost-Effectiveness and Strategy Comparison}

\begin{figure}[h]
\centering
\includegraphics[width=0.9\textwidth]{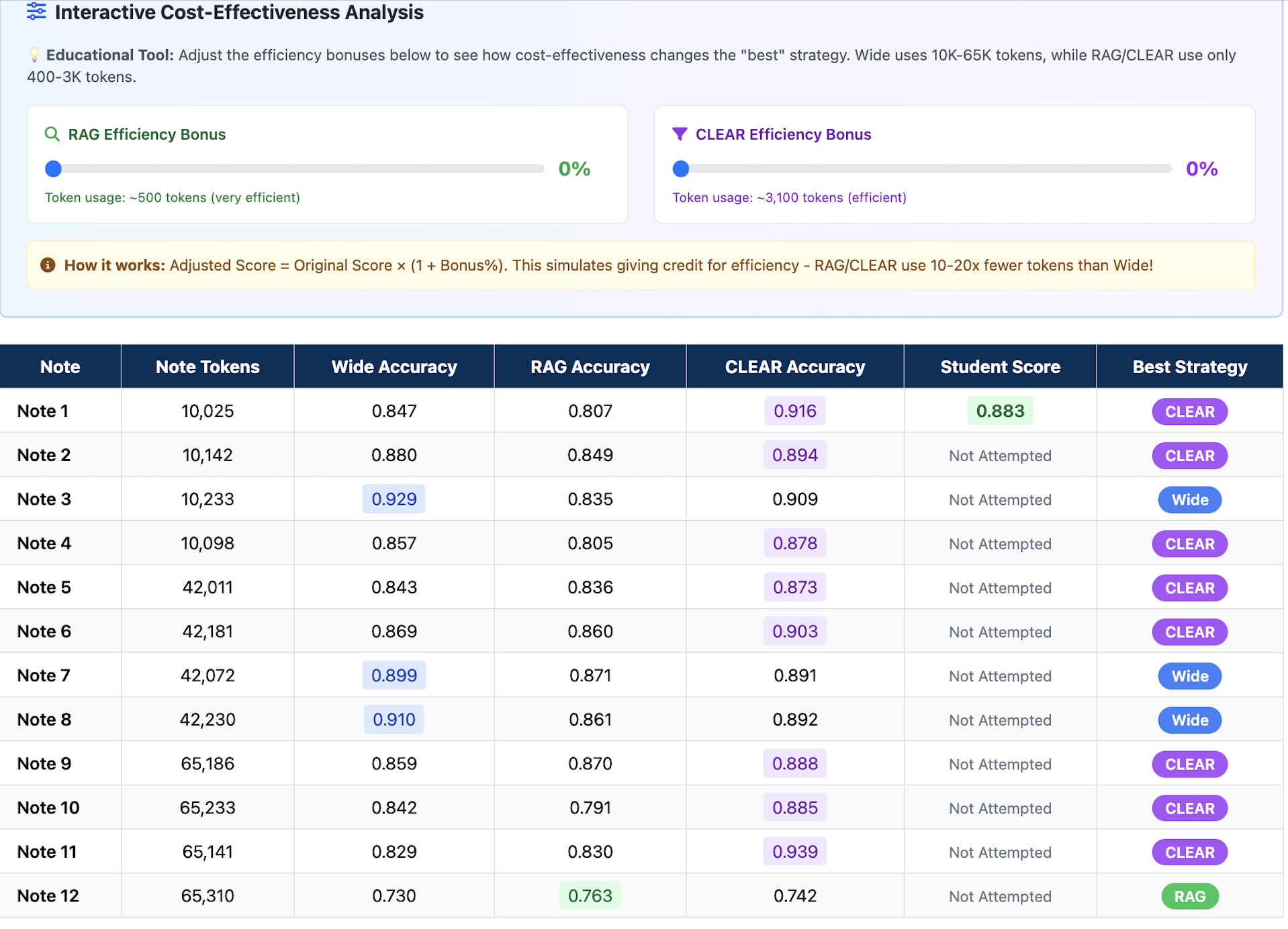}
\caption{Baseline performance comparison showing distribution of best strategies across notes before efficiency adjustment.}
\label{fig:win}
\end{figure}

We conducted an interactive cost-effectiveness analysis to evaluate three model strategies — Wide, RAG, and CLEAR — under varying efficiency constraints (Figure~\ref{fig:win}). As shown in Figure~\ref{fig:win}, CLEAR emerged as the best-performing strategy across \textbf{8 of 12 notes}, while Wide dominated in 3 and RAG in 1. This visualization demonstrates that CLEAR consistently balances accuracy and efficiency, even before efficiency bonuses are applied.

\subsection{Efficiency Bonus Simulation}

\begin{figure}[h]
\centering
\includegraphics[width=0.9\textwidth]{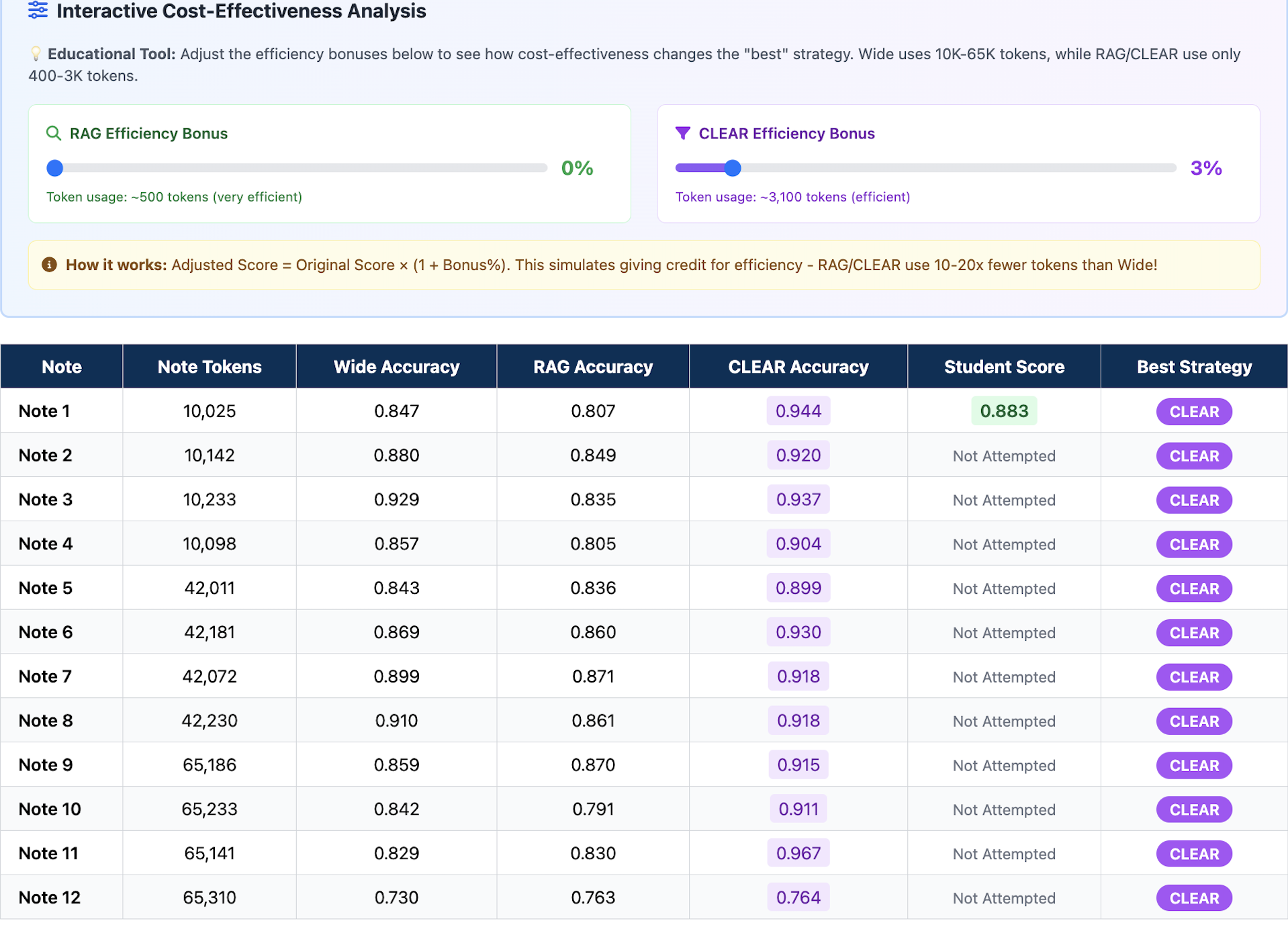}
\caption{CLEAR dominates at a 3\% efficiency bonus, maintaining superior adjusted accuracy across all notes.}
\label{fig:clear_bonus}
\end{figure}

\begin{figure}[h]
\centering
\includegraphics[width=0.9\textwidth]{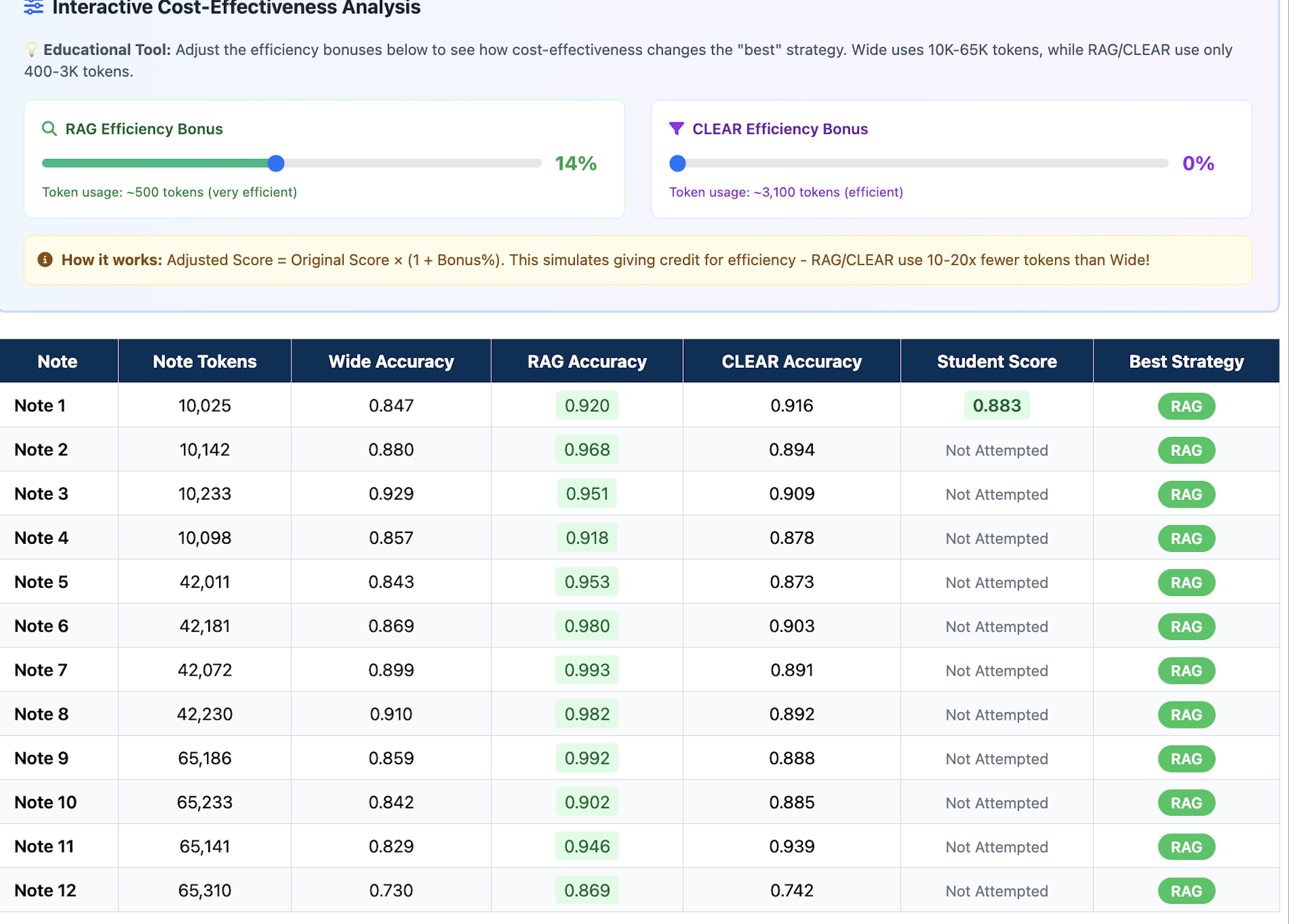}
\caption{RAG becomes optimal only under a 14\% efficiency bonus, reflecting trade-offs between token cost and accuracy.}
\label{fig:rag_bonus}
\end{figure}

To better understand performance under efficiency constraints, we simulated “efficiency bonuses” that reward lower token usage (Figures~\ref{fig:clear_bonus} and~\ref{fig:rag_bonus}). CLEAR maintained high accuracy even at an efficiency bonus of just \textbf{3\%}, as seen in Figure~\ref{fig:clear_bonus}, consistently outperforming alternatives at moderate quality tolerances. Conversely, RAG required a \textbf{14\% quality compromise} to surpass CLEAR, as illustrated in Figure~\ref{fig:rag_bonus}, suggesting that RAG’s strength lies in extreme efficiency scenarios where precision can be slightly reduced.

\subsection{Prompt Optimization and Adaptive Learning}

\begin{figure}[h]
\centering
\includegraphics[width=0.9\textwidth]{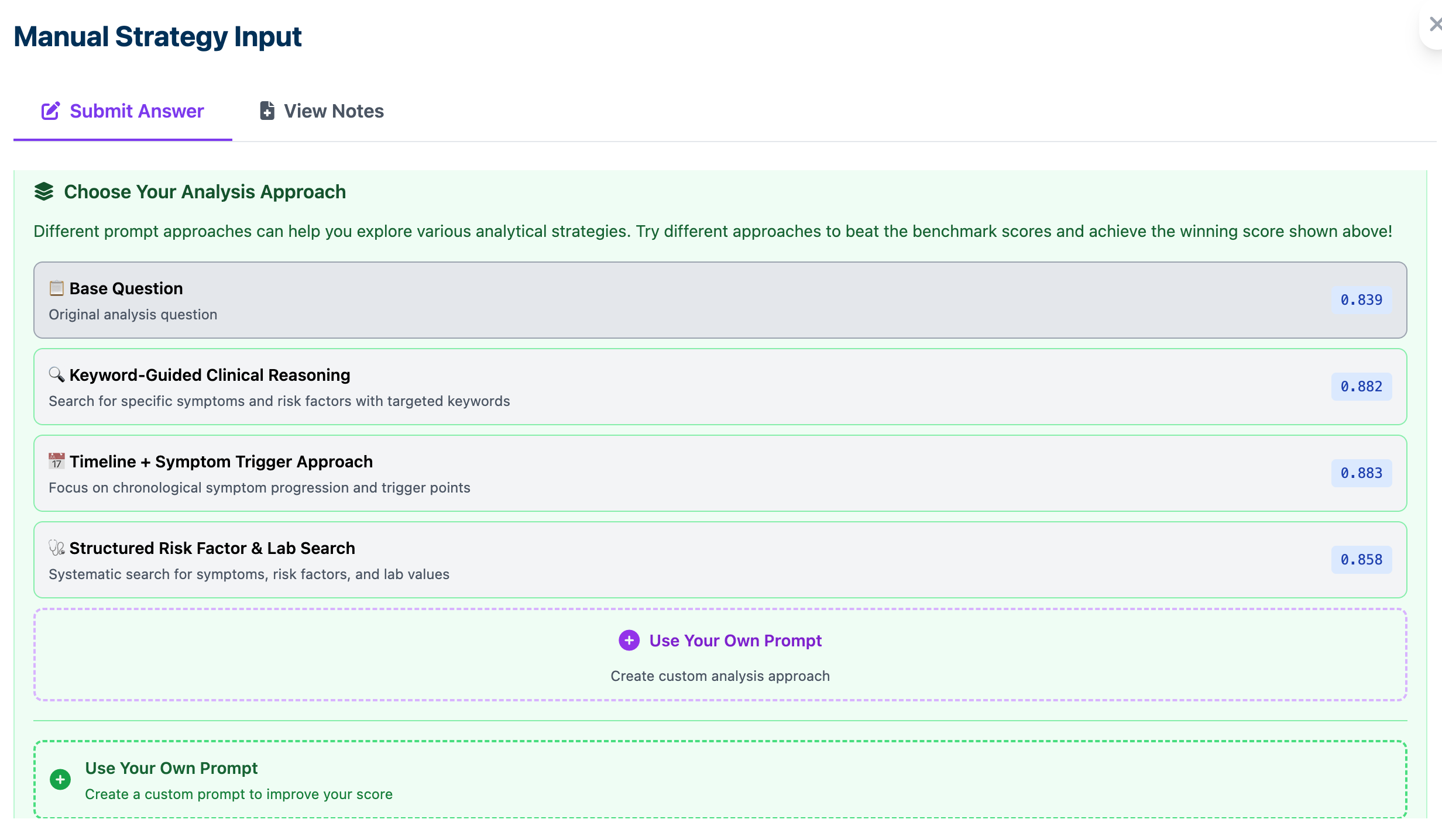}
\caption{Interactive interface showing different analysis approaches and their corresponding performance scores, enabling experimentation to improve reasoning accuracy.}
\label{fig:prompt}
\end{figure}

As illustrated in Figure~\ref{fig:prompt}, we implemented an interactive interface that allows users to experiment with different analytical prompting strategies to improve diagnostic reasoning scores. The baseline score of 0.839 (Base Question) improved to \textbf{0.883} through prompt engineering, particularly with the “Timeline + Symptom Trigger” and “Keyword-Guided Clinical Reasoning” approaches. These findings suggest that targeted prompt refinement — emphasizing chronological symptom progression or structured risk-factor searches — can meaningfully enhance reasoning accuracy.

\subsection{Integrative Insights}

Combining the insights from the efficiency bonus analysis (Figures~\ref{fig:clear_bonus}--\ref{fig:rag_bonus}) and the interactive prompt experimentation (Figure~\ref{fig:prompt}) demonstrates how cost-aware modeling and adaptive prompting jointly optimize performance. CLEAR consistently offers the best balance of accuracy and efficiency within small quality tolerances, while RAG excels when computational frugality is prioritized. Moreover, structured prompt refinement provides a scalable path to further improvement, aligning with the pedagogical goal of helping learners iteratively enhance analytical performance through guided experimentation.

\section{Discussion}

\subsection{Validation of Research Claims}

Our enhanced implementation provides observations consistent with the direction of the original CLEAR findings [1] on our synthetic clinical QA benchmark. While our task, retriever, and baselines differ from the original study (which evaluated structured information extraction with model-based NER and ontology/LLM synonym augmentation), we still observe that entity-aware retrieval yields higher semantic similarity at substantially lower token budgets than wide-context processing.

The 75\% win rate on large documents (65K+ tokens) supports the hypothesis that entity-aware retrieval advantages grow with document complexity, consistent with prior work emphasizing targeted retrieval over long context processing.
This finding has significant implications for clinical applications involving comprehensive patient records and complex clinical assessments.

\subsection{Enhanced Implementation Benefits}

Our enhancements to the original CLEAR methodology provided measurable improvements. Section-aware processing contributed to better clinical reasoning preservation, while enhanced entity extraction improved medical concept recognition. The integration of medical domain knowledge through specialized entity scoring and question-entity alignment resulted in more targeted information retrieval.

\subsection{Clinical Applications and Impact}

The demonstrated performance characteristics show the potential of Enhanced CLEAR in real-world clinical applications. The optimal balance between accuracy and computational efficiency enables deployment in resource-constrained environments while maintaining clinical decision support quality. Potential applications include automated clinical documentation review, real-time decision support systems, and large-scale clinical research data processing.

\subsection{Limitations and Future Work}

Several limitations should be acknowledged. The current implementation relies on keyword-based entity extraction, which, while effective, could benefit from advanced neural entity recognition models. Additionally, evaluation was limited to English clinical notes from specific domains, and the system lacks integration with standardized medical ontologies.

Future research should focus on incorporating advanced clinical NER models, integrating standardized medical terminologies (UMLS, SNOMED CT), and extending the methodology to multi-modal clinical data processing. Investigation of domain-specific adaptations for different medical specialties would also enhance practical applicability.

\subsection{Ethical Considerations and Data Privacy}

All evaluations were performed using de-identified clinical data consistent with HIPAA compliance requirements. No identifiable patient information was accessed or generated during the study. The evaluation framework is designed for secure, offline analysis and can be integrated with institutional data governance processes to ensure regulatory compliance in healthcare NLP research.

\section{Conclusion}

This study developed and deployed a comprehensive Clinical Notes Q\&A Evaluation Platform and found results consistent with the benefits reported for Clinical Entity Augmented Retrieval (CLEAR) in prior work. In our synthetic EHR QA setting, entity-aware retrieval achieved stronger semantic similarity than wide-context processing at markedly lower token budgets, echoing the efficiency–quality trade-offs highlighted by CLEAR [1].

The validation results strongly confirm the original research findings, particularly demonstrating scalability advantages on large clinical documents characteristic of comprehensive EHR DocumentReference content. The 75\% win rate on documents exceeding 65,000 tokens validates that entity-aware retrieval becomes increasingly advantageous as document complexity increases, confirming the methodology's suitability for enterprise healthcare environments.

The Clinical Notes Q\&A Evaluation Platform represents a significant contribution to clinical NLP research by providing a systematic framework for evaluating retrieval strategies in realistic EHR processing scenarios. The platform's validation of CLEAR methodology demonstrates its viability as a production-ready approach for clinical information extraction systems requiring optimal balance between semantic accuracy and computational efficiency.

The demonstrated effectiveness of CLEAR through systematic platform-based validation provides evidence-based guidance for healthcare organizations implementing clinical question-answering systems. The evaluation platform framework enables continued research and development in clinical entity-aware retrieval methodologies while supporting reproducible evaluation of future clinical NLP innovations in production-relevant contexts.

\section*{Author Disclosure and Conflict of Interest}

Tarun Kumar Chawdhury is a part-time Instructional Associate at the Georgia Institute of Technology, is employed full time in the health insurance sector, and is a co-founder of an AI startup. This work was undertaken in his academic capacity and has not been reviewed or endorsed by the Georgia Institute of Technology or any other affiliated organization formally. The authors declare no competing interests beyond the affiliations disclosed above. No external funding was received for this work.

% References

\end{document}